\pdfoutput=1
\documentclass[11pt]{article}

\usepackage[]{acl}

\usepackage{times}
\usepackage{latexsym}

\usepackage[T1]{fontenc}
\usepackage[utf8]{inputenc}

\usepackage[protrusion=false,expansion=true,kerning=true,babel=true,final]{microtype}

\usepackage{amssymb}
\usepackage{bbm}
\usepackage{bm}
\usepackage{graphicx}
\usepackage{caption}
\usepackage{adjustbox}
\usepackage{subcaption}
\usepackage{amsmath}
\usepackage{multirow}
\usepackage{makecell}
\usepackage{booktabs}
\usepackage{tcolorbox}
\usepackage[inline]{enumitem}
\usepackage{float}
\usepackage{hyperref}
\usepackage{float}
\usepackage[title]{appendix}

\usepackage{titlesec}
\titlespacing*{\section}{0pt}{1ex plus .2ex minus .2ex}{0.6ex plus .2ex minus .3ex}
\DeclareMathOperator*{\argmax}{arg\;max}
\newcommand{\outsider}{\textit{Outsider}}
\newcommand{\insider}{\textit{Insider}}
\newcommand{\post}{\mathtt{p}}
\newcommand{\noun}{\mathtt{n}}
\newcommand{\dataset}{\mathcal{D}}
\newcommand{\token}{t}
\newcommand{\loss}{\mathcal{L}}
\newcommand{\pred}{\bm{\pi}}

\title{Which side are you on? Insider-Outsider classification in conspiracy-theoretic social media}
\author{Pavan Holur{\normalfont \textsuperscript{1}}, 
        Tianyi Wang{\normalfont \textsuperscript{1}}, 
        Shadi Shahsavari{\normalfont \textsuperscript{1}}, \\ \textbf{Timothy Tangherlini}{\normalfont \textsuperscript{2}}, \and \textbf{Vwani Roychowdhury}{\normalfont \textsuperscript{1}} \\
        {\normalfont \textsuperscript{1}} Department of Electrical and Computer Engineering, UCLA \\
        {\normalfont \textsuperscript{2}} Department of Scandinavian, UC Berkeley \\
        \texttt{\{pholur,tianyiw,shadihpp,vwani\}@ucla.edu, tango@berkeley.edu} }
\begin{document}
\setlength{\belowdisplayskip}{0.5em} \setlength{\belowdisplayshortskip}{0.5em}

\maketitle

\begin{abstract}

Social media is a breeding ground for threat narratives and related conspiracy theories. In these, an \textit{outside} group threatens the integrity of an \textit{inside} group, leading to the emergence of sharply defined group identities: \insider{}s -- agents with whom the authors identify and \outsider{}s -- agents who threaten the insiders. Inferring the members of these groups constitutes a challenging new NLP task: (i) Information is  distributed over many poorly-constructed posts; (ii) Threats and threat agents are highly contextual, with the same post potentially having multiple agents assigned to membership in either group; (iii) An agent's identity is often implicit and transitive;  and (iv) Phrases used to imply \outsider{} status often do not follow common negative sentiment patterns. To address these challenges, we define a novel \insider{}-\outsider{}  classification task. Because we are not aware of any appropriate existing datasets or attendant models, we introduce a labeled dataset (CT5K) and design a model (NP2IO) to address this task. NP2IO leverages pretrained language modeling to classify  \insider{}s and \outsider{}s.
% We compare the performance of this model to other candidate models such as NB and CBOW.
NP2IO is shown to be robust, generalizing to noun phrases not seen during training, and exceeding the performance of non-trivial baseline models by $20\%$.

\end{abstract}

\begin{figure*}
    \centering
    \includegraphics[width=0.90\textwidth]{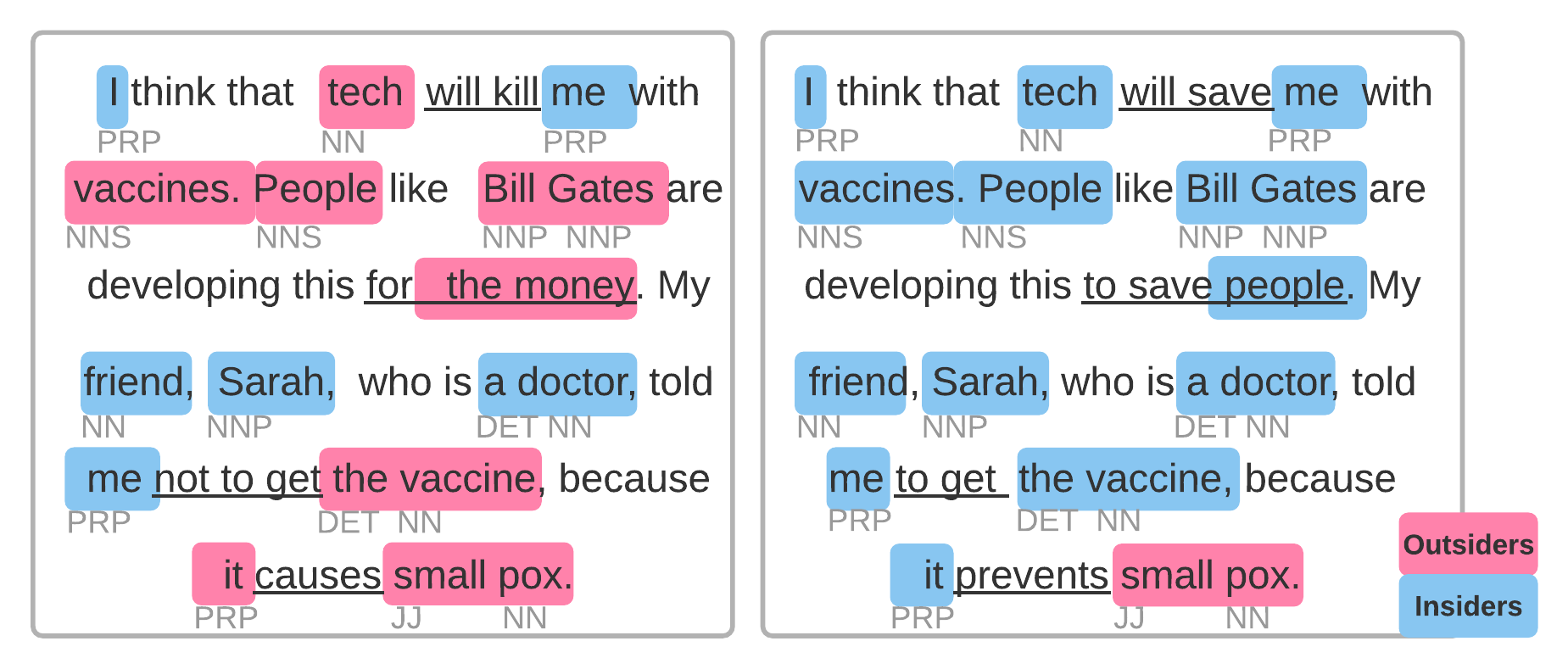}
    % \caption{\textbf{An inferred post by NP2IO:} The phrases in red represent the perceived outsiders in the post as determined by the post's author and the phrases in blue represent the insiders. In this post, \#BigTech is an \textit{Outsider} while doctors are \textit{Insiders}.}
    \caption{\textbf{A pair of inferred text segments labeled by NP2IO showing \textit{Insider-Outsider} context-sensitivity:} Colored spans are used to highlight noun phrases that are inferred (red for \outsider{}s; blue for \insider{}s). POS tags are shown along with the noun phrases to illustrate an example of syntactic and semantic hints used by NP2IO to generate the inferred labels. Note that, based solely on context, the same agents (``tech'', ``vaccines'', ``People'', ``Bill Gates'' and ``the vaccine'') switch \textit{Insider-Outsider} label. \textit{Even though the training data is highly biased in terms of the identities of the \insider{}s/\outsider{}s, the pretrained language model used in our classifier allows NP2IO to learn to infer using the context phrases and} not \textit{by memorizing the labels.}}
    \label{fig:intro}
\end{figure*}

% 1. Narratives are important
%    - Influence to real world
%    - Reflection of people's view of the real world
% 2. Conspiracies has always been particularly important
%    - ? They are driven by real events, but then take a leap across a gap between people's belief/ideology and the reality, highlighting the characteristics of different groups/communities and their separation/conflicts.
% 3. Social media boosts the development of narratives more than ever. On one hand it leads to abundant text for analysis; on the other hand it also empowers the unprecedented real-world effects of narratives, especially conspiracies. Therefore, it is the best-ever time for analysis of conspiracy theoretic narratives.
% 4. The analysis methodology of conspiracy theoretic narratives: narrative theories.
%    - Essential elements of a conspiracy theory: threat, actant/perpetrator (outsider), victim (insider), (optional) strategy
%.   - @Pavan NEED HELP: the identification of insiders and outsiders is the most important part of the job??? Or why do we focus on this? (besides it's more feasible of a task) - @Tim I am hoping that we can motivate this with all the narrative theory in a relevant section.@Tianyi... the green highlight is where I think this might make sense. IDK.

\section{Background and Motivation}
% \iffalse The concept of ``narrative'' has emerged as a leading abstract and a succinct model that aims to capture how seemingly different behavior patterns, policies and identities emerge at multiple levels of the societal hierarchy -- ranging from the individual to nationhood and beyond. For example, computational linguists believe that authors write to capture and reflect on dominant narratives, which are therefore considered the generative backbone of stories \cite{bailey, good_for}. This underlying narrative structure, common to all stories, provides a template to the story generation model, specifying themes, key plotlines, familiar archetypes, the heroes and villains, protagonists, antagonists and the ebbs and flows of emotions. One well-established model of the narrative is a narrative network of characters, their roles, their interactions (syuzhet) and associated time-sequencing information (fabula). Thus, stories can be considered as instantiations of an underlying narrative framework.  Much research in Automatic Story Generation (ASG) \cite{asg, asg_2} attempts to both learn such time-sequenced network models, as well as, write stories by instantiating them.
% \fi

Narrative models -- often succinctly represented as a network of characters, their roles, their interactions (\textit{syuzhet}) and associated time-sequencing information (\textit{fabula}) -- have been  a subject of considerable interest in computational linguistics and narrative theory.  \iffalse It is hypothesized that  narratives are the generative backbone of stories\fi Stories rest on the generative backbone of narrative frameworks \cite{bailey, good_for}. While the details \iffalse (such as who, where, when, and the setting of the plot)\fi might vary from one story to another, this variation can be compressed into a limited set of domain-dependent narrative roles and functions \cite{dundes1962morphology}. 
% Research into Automatic Story Generation (ASG) \cite{asg, asg_2} propose various methods for learning time-sequenced network models of narratives as well as writing stories by instantiating these models.
% \iffalse  
% It has been established by  scholars from different fields studying human psychology, sociology, folklore and legends have convincingly argued that narratives are an inherent way for humans to extract and represent meaning both internally and externally. In fact, stories -- driven by underlying master narratives -- are everywhere: we employ them to convey our personal experiences, our opinions about real-world events, in order to contextualize our beliefs and ultimately to define our identities. As opposed to literary works, such endemic narratives are seldom written down and are themselves constantly evolving, driven by a collective process of negotiations and the urgency demanded by external events and threats. Such dynamics and upheavals notwithstanding, retrospective studies have shown that these negotiated narratives at any particular time have the same overall structure as the ones used to model the more structured world of classical literature. 
% \fi 

% \iffalse As opposed to the literary work scenario,  s\fi
Social narratives that both directly and indirectly contribute to the construction of
% \iffalse define identities and policies of individuals and groups\fi
individual and group identities are an emergent phenomenon resulting from
distributed social discourse. Currently, this phenomenon is most readily
apparent on social media platforms, with their large piazzas and niche enclaves.
Here, multiple threat-centric narratives emerge and, often, over time are
\iffalse labeled as\fi linked together into complex conspiracy theories
\cite{tangherlini_2}. Conspiracy theories, and their constituent threat
narratives (legend, rumor, personal experience narrative) share a signature
semantic structure: an implicitly accepted \insider{} group; a diverse group of
threatening \outsider{}s; specific threats from the \outsider{} directed at the
\insider{}s; details of how and why \outsider{}s are threatening; and a set of
strategies proposed for the \insider{}s to counter these threats
\cite{tangherlini2018toward}. Indeed, the \insider{}/\outsider{}
groups are fundamental in most studies of belief narrative, and have been exhaustively studied in social theory and more specifically, in the context of conspiracy
theories \cite{redo-bodner2020covid,Barkun+2013}. 
% - \textit{we form our beliefs about narratives and their validity, in order to alter our perception of the world to improve our feeling of well-being, despite the fact that in most cases these beliefs are based on dubious and/or empirically false evidence}. 
On social media, these narratives are negotiated one post at a time,
expressing only short pieces of the ``immanent  narrative whole''
\cite{clover1986long}. This gives rise to a new type of computational linguistic
problem: \textit{Given a large enough corpus of social media text
data, can one automatically distill  semantically-labeled narratives
(potentially several overlapping ones) that underlie the fragmentary
conversational threads?}

Recent work \cite{tangherlini_1, tangherlini_2, shadi_0, rsos} has shown considerable promise that such scalable automated algorithms can be designed.
% \iffalse They have extracted  overlapping narratives from conspiracy theoretic social media data, comprising posts from 4chan, 8kun, and Reddit as well as comments on YouTube. , posts on Reddit and submissions to 8kun.\fi
An automated pipeline of interlocking machine learning modules decomposes the posts into actors, actants and their inter-actant relationships to create narrative networks via aggregation.
% \iffalse consists of state-of-the-art submodules such as open information extraction (OpenIE), wordpiece embeddings (BERT) and clustering algorithms (HDBSCAN). This pipeline\fi
\textit{These network representations are interpretable on \iffalse manual\fi inspection}, allowing for the easy identification of the various signature semantic structures: %the protagonists, antagonists, perpetrators, agitators, threats, 
\insider{}s, \outsider{}s, strategies for dealing with \outsider{}s and their attendant threats and, in the case of conspiracy theories, causal chains of events that support that theory.

\textit{By itself, this unsupervised platform does not ``understand'' the different narrative parts}. Since the submodules are not trained to look for specific semantic abstractions inherent in conspiracy theories, the platform cannot automatically generate a semantically tagged narrative for downstream NLP tasks. It cannot, for example, generate a list across narratives of the various outside threats and attendant inside strategies being recommended on a social media forum, nor can it address why these threats and strategies are being discussed. 
%THIS DANGLES: Accomplishing abstract tasks such as identifying what constitutes a threat or an agent of threat requires some form of supervision incorporating real-world knowledge.

\section{The Novel Insider vs. Outsider Classification Problem}\label{io}
As a fundamental first step bringing in supervised information to enable automated narrative structure discovery, we introduce the \insider{}-\outsider{} classification task: To classify the noun phrases in a post as \insider{}, \outsider{} or \textit{neither}.

A working conceptualization of what we consider \insider{}s and \outsider{}s is provided in the following insets. As with most NLP tasks, we do not provide formal definitions of and rules to determine these groups. Instead we let a deep learning model learn the  representations needed to capture these notions computationally by training on data annotated with human-generated labels. 
%A first-order task is to identify the foundational archetypes of narratives:  who are the heroes and the villains in this narrative, or more generally, what are the \insider{}s and \outsider{}s.

% \begin{tcolorbox}
% \begin{description}[leftmargin=0pt,itemindent=0pt,style=sameline]
% \item[Insiders:] Some combination of actors and their associated pronouns, with full agency (people, organizations, government), partial agency (policies, laws, rules, current events) or no agency (things, places, circumstances), that the author sides with, including themselves. These are often ascribed beneficial status;

% \item[Outsiders:] A set of actors whom the author opposes and in many cases perceives as a set of agents threatening the author and the group of insiders with harm. These agents need not have full agency according to the author of the post: Diseases, natural disasters, for example, would be universal outsiders.
% \end{description}
% \end{tcolorbox}
\begin{tcolorbox}[float=h!,title=Conceptualization of \insider{}s and \outsider{}s,enlarge bottom finally by=-0.5em]
\begin{description}[leftmargin=0pt,itemindent=0pt,style=sameline]
\item[Insiders:] Some combination of actors and their associated pronouns, who display full agency (people, organizations, government), partial agency (policies, laws, rules, current events) or no agency (things, places, circumstances), with whom the author identifies (including themselves). These are often ascribed beneficial status;
% \end{description}
% \end{tcolorbox}

% \begin{tcolorbox}[float=htpb!]
% \begin{description}[leftmargin=0pt,itemindent=0pt,style=sameline]
\item[Outsiders:] A set of actors whom the author opposes and, in many cases, perceives as threatening the author and the insiders with disruption or harm. For our purposes, \textit{these agents need not have full agency}: %according to the author of the post: 
Diseases and natural disasters, for example, would be universal outsiders, and any man-made object/policy that works against the \insider{}s would be included in this group.
\end{description}
%\vspace{-2em}
\end{tcolorbox}

The partitioning of actors from a post into these different categories \iffalse resembles closely, and\fi is inspired by social categorization, identification and comparison in the well-established Social Identity Theory (SIT) \cite{tajfel_1, tajfel_2} and rests on established perspectives from Narrative Theory \cite{dundes1962morphology,labov1967narrative,nicolaisen1987linguistic}. 

Following are some of the reasons why this classification task is challenging and why the concepts of \insider{}s/\outsider{}s are not sufficiently captured by existing labeled datasets used in Sentiment Analysis (SA) (discussed in more detail in Section \ref{related}):

\begin{enumerate}[leftmargin=0em,topsep=0em minus 0.5em,itemsep=0em, wide, labelindent=0pt%, listparindent=0pt%
]
  
\item \textbf{\iffalse Narrative-Specific \fi Commonly-held Beliefs and Worldviews:} Comprehensively incorporating  shared values, crucial to the classification of \insider{}s and \outsider{}s, is a task with varied complexity. Some beliefs are easily enumerated: most humans share a perception of a nearly universal set of threats (virus, bomb, cancer, \iffalse autism,\fi dictatorship) or threatening actions (``kills millions of people'', ``tries to mind-control everyone'') or benevolent actions (``donating to a charitable cause'', ``curing disease'', ``freeing people''). Similarly, humans perceive themselves and their close family units as close, homogeneous groups with shared values, and therefore  ``I'', ``us'', ``my children'' and ``my \iffalse spouse\fi family'' are usually \insider{}s. In contrast, ``they'' and ``them'' are most often \outsider{}s.

Abstract beliefs pose a greater challenge as the actions that encode them can be varied and subtle.   For example, in the post: ``The microchips in vaccines track us'', the noun phrase ``microchips'' is in the \outsider{} category as \iffalse any means for surveillance\fi it violates \iffalse is assumed to be against \fi the \insider{}s' right to privacy by ``track[ing] us''. Thus, greater attention needs to be paid in labeling datasets, highlighting ideas such as the right to freedom, religious beliefs, and notions of equality. 

\iffalse
. The clues \iffalse for the classification of \fi to classify an agent as an \insider{} or \outsider{} can at times be explicitly stated within a social media post or universally recognized as such. For example, sources of objective hazard (virus, bomb, cancer, autism, dictatorship), threatening actions (``kills millions of people'', ``tries to mind-control everyone'') or direct description (``is evil'') are definite evidence indicating \outsider{}s, and can be potentially identified by existing labeled datasets. On the other hand, the  first person ``I'' in a post, would almost always be an \insider{}, and most likely so are ``my children'' and ``my spouse''. 

In many cases, however, the clues in a post need to be contextualized within the overarching narrative structure. In general, noun phrases representing beneficial categories such as ``children'', ``family'' and beliefs such as ``right to privacy'', ``freedom of choice'' and ``good health'' are often implicitly assumed as \insider{}s in a narrative. As a result, in a post that says, ``The microchips in vaccines track us,'' the noun phrase ``microchips'' is in the \outsider{} category as any means for surveillance violates \iffalse is assumed to be against \fi the \insider{}s' rights.   
\fi

\item \textbf{Contextuality and Transitivity:} People express their opinions of \textit{Insider/Outsider} affiliation by adding \textit{contextual} clues  that are embedded in the language of social media posts. For example, a post ``We should build cell phone towers'' suggests that ``cell phone towers'' are helpful to \insider{}s, whereas a post ``We should build cell phone towers and show people how it fries their brains'' suggests, in contrast, that ``cell phone towers'' are harmful to \textit{Insiders} and belong, therefore, to the class of \outsider{}s. \textit{Insider/Outsider} affiliations are also implied in a \textit{transitive} fashion within a post. For example, consider two posts: (i)  ``Bill Gates is developing a vaccine. Vaccines \textit{kill} people.'' and (ii) ``Bill Gates is developing a vaccine. Vaccines \textit{can eradicate} the pandemic.'' In the first case, the vaccine's toxic quality and attendant \outsider{} status would transfer to Bill Gates, making him an \outsider{} as well; in the second post, vaccine's beneficial qualities would transfer to him, now making ``Bill Gates'' an \insider{}. 

\item \textbf{Model Requirement under Biased Data Conditions:} Designing effective classifiers that do not inherit bias from the training data -- especially data in which particular groups or individuals are derided or dehumanized --  is a challenging but necessary task. Because conspiracy theories evolve, building on earlier versions, and result in certain communities and individuals being ``othered'', our  models \textit{must} learn the phrases, contexts, and transitivity used to ascribe group membership, here either \insider{}s or \outsider{}s and not memorize the communities and/or individuals being targeted. Figure~\ref{fig:intro} illustrates an example where we probed our model to explore whether such a requirement is indeed satisfied. The first text conforms to the bias in our data, where ``tech'', ``Bill Gates'', and ``vaccines'' are primarily \textit{Outsiders}. The second text switches the context by changing the phrases. Our classifier is able to correctly label these same entities, now presented in a different context, as \insider{}s! We believe that such subtle learning is possible because of the use of pretrained language models. We provide several such examples in Table~\ref{tab:adv} and Figure~\ref{fig:examples} and also evaluate our model for Zero-shot learning in Table~\ref{tab:full_res} and Figure \ref{fig:heatmap1}.  

\end{enumerate}

\section{Our Framework and Related Work}\label{related}

% FLIP IT AROUND MAYBE MOTIVATE FROM INSIDER OUTSIDER - NOT FORMING GROUPS - approve and disapprove in a utilitarian sense. Span is the whole sentence
% whether in our language whether the threat is the insider span is the whole

Recent NLP efforts have examined the effectiveness of using pretrained Language Models (LM) such as BERT, DistilBERT, RoBERTa, and XLM to address downstream classification tasks through fine-tuning \cite{distilbert, roberta, xlm}. Pretraining establishes the contextual dependencies of language prior to addressing a more specialized task, enabling rapid and efficient transfer learning. A crucial benefit of pretraining is that, in comparison to training a model from scratch, fewer labeled samples are necessary. By fine-tuning a pretrained LM, one can subsequently achieve competitive or better performance on an NLP task.  As discussed in Section~\ref{io}, since our model is required to be \textit{contextual} and \textit{transitive}, both of which are qualities that rely on the context embedded in language, we utilize a similar architecture. 

In recent work involving span-based classification tasks, token-classification heads have proven to be very useful for tasks such as, Parts-of-Speech (POS) Tagging, Named Entity Recognition (NER) and variations of Sentiment Analysis (SA) \cite{emotionxku, bertbi, sentibert}. Since the \insider{}-\outsider{} classification task is also set up as a noun phrase labeling task, our architecture uses a similar token-classification head on top of the pretrained LM backbone.
%As \insider{}-\outsider{} classification is set up as a noun phrase labeling task in the larger context of a social media post, our BERT architecture drives a conventional token-classification head. These classification heads are proven to be useful in a myriad of span-based classification tasks including Parts-of-Speech (POS) Tagging, Named Entity Recognition (NER) and variations of Sentiment Analysis (SA) \cite{emotionxku}, \cite{bertbi}, \cite{sentibert}. 

%
Current SA datasets' definitions of positive negative and neutral sentiments can be thought of as a ``particularized'' form of the \textit{Insider-Outsider} classification task. For example, among the popular datasets used for SA, Rotten Tomatoes, Yelp reviews \cite{sst} and others \cite{dong-etal-2014-adaptive, pontiki-etal-2014-semeval} implicitly associate a sentiment's origin to the post's author (source) (a single \insider{}) and its intended target to a movie or restaurant (a single \outsider{} if the sentiment is \textit{negative} or an \insider{} if \textit{positive}). The post itself generally contains information about the target and particular aspects that the \insider{} found necessary to highlight. 

In more recent SA work, such as Aspect-Based Sentiment Analysis (ABSA) \cite{aspect, aspect_2, wang, absa_rel},  researchers have developed models to extract sentiments -- positive, negative, neutral -- associated with particular aspects of a target \textit{entity}. One of the subtasks of ABSA, aspect-level sentiment classification (ALSC), has a form that is particularly close to the \insider{}-\outsider{} classification. Interpreted in the context of our task, the author of the post is an \insider{} although now there can potentially be multiple targets or ``aspects'' that need to be classified as \insider{}s and \outsider{}s. Still, the constructed tasks in ABSA appear to not align well with the goal of \insider{}-\outsider{} classification:
\begin{enumerate*}[label=\arabic*)]
    % \item Most of the tasks attempt to classify only NER-tagged actors rather than all noun phrases, which may fail when posts are poorly capitalized and punctuated; % For the ALSC task, the noun phrases are not limited to NER, while for the aspect extraction subtask of ABSA, the goal is explicitly the aspects, which differs from NER's goal, too
    \item Datasets are not \textit{transitive}: Individual posts appear to have only one agent that needs classification, or a set of agents, each with their own separate sets of descriptors;
    \item The ALSC data is often at the sentence-level as opposed to post-level, limiting the context-space for inference.
\end{enumerate*}
Despite these obvious differences, we quantitatively verify our intuitions in Section~\ref{tianyi_og}, and show that ABSA models do not generalize to our dataset.

Closely related to ABSA is Stance Classification (SC) (also known as Stance Detection / Identification), the task of identifying the stance of the text author (\texttt{in favor of}, \texttt{against} or\; \texttt{neutral}) toward a target (an entity, concept, event, idea, opinion, claim, topic, etc.)\cite{walkerStanceClassificationUsing2012, zhangWeMakeChoices2017, kucukStanceDetectionSurvey2021}. Unlike ABSA, the target in SC does not need to be embedded as a span within the context. For example, a perfect SC model given an input for classification of context: \textit{This house would abolish the monarchy.} and target: \textit{Hereditary succession}, would predict the \textit{Negative} label \cite{redone1, redone2}. While SC appears to require a higher level of abstraction and, as a result, a model of higher complexity and better generalization power than those typically used for ABSA, current implementations of SC are limited by the finite set of queried targets; in other words, SC models currently do not generalize to unseen abstract targets. Yet, in real-time social media, potential targets and agents exhibit a continuous process of emergence, combination and dissipation. We seek to classify these shifting targets using the transitive property of language, and would like the language to provide clues about the class of one span \textit{relative} to another. Ultimately, while SC models are a valuable step in the direction of better semantic understanding, they are ill-suited to our task.

Parallel to this work in SA, there are complementary efforts in consensus threat detection on social media \cite{wester, kandias, park}, a task that broadly attempts to classify longer segments of text -- such as comments on YouTube or tweets on Twitter -- as more general ``threats''. The nuanced instruction to the labelers of the data is to \textit{identify whether the author of the post is an} \outsider{} \textit{from the labeler's perspective as an } \insider{}. Once again, we observe that this task aligns with the \insider{}-\outsider{} paradigm, but does not exhaust it, and the underlying models cannot accomplish our task.

The sets of \insider{}s and \outsider{}s comprise a higher-order belief system that cannot be adequately captured with the current working definitions of sentiment nor the currently available datasets. This problem presents \textit{a primary motivation for creating a new dataset}. For example, the post: ``Microchips are telling the government where we are'', does not directly feature a form of prototypical sentiment associated with ``microchips'', ``the government'' and ``we'', yet clearly insinuates an invasion on our right to privacy making clear the \insider{}s (``we'') and \outsider{}s (``microchips'', ``the government'') in the post.

\section{Data Collection}

% We construct a dataset that conforms to the three requirements specified in Section \ref{io}:
% \begin{itemize}
%     \item \textbf{Representative of an Underlying Narrative:} The posts in the dataset must be collected as samples of an underlying narrative that is socially relevant and has a large footprint in conspiracy theoretic circles.
%     \item \textbf{Sufficiently Rich:} The dataset must possess enough samples in total to capture labels for commonly-held beliefs and threats and enough samples per popular agent to specify the patterns of Transitivity and Contextuality required by the downstream classification model.
% \end{itemize}
% \noindent \textbf{Approach:} 
To construct our novel dataset -- \textbf{C}onspiracy \textbf{T}heory-5000 (\textbf{CT5K}) -- we designed crawlers to extract a corpus of social media posts generated by the underlying narrative framework of vaccine hesitancy (Details of the crawlers are documented in Appendix \ref{app:data_collection}). Vaccine hesitancy is a remarkably resilient belief fueled by conspiracy theories that overlaps with multiple other narratives including ones addressing ``depopulation'', ``government overreach and the deep state'', ``limits on freedom of choice'' and ``Satanism''. The belief's evolution on social media has already enabled researchers to take the first steps in modeling critical parts of the underlying generative models that drive anti-vaccination conversations on the internet \cite{mommy, roja}. Moreover, vaccine hesitancy is especially relevant in the context of the ongoing COVID-19 pandemic \cite{burki}. 

On the crawled corpus, we extract the noun-chunks from each post using SpaCy's noun chunk extraction module and dependency parsers \cite{spacy_OG}. A noun chunk is a sub-tree of the dependency parse tree, the headword of which is a noun. The result is a set of post-phrase pairs, $(\post, \noun)$, where $\post$ is a post and $\noun$ is one of the noun phrases extracted from the post.

Amazon Mechanical Turk (AMT) (see Appendix \ref{app:AMT} for labeler instructions) was used to label the post-phrase pairs. For each pair, the labeler was asked, \textit{given the context}, whether the writer of the post $\post$ perceives the noun phrase $\noun$ to be an \insider{}, \outsider{} or \textit{neither} (N/A). The labeler then provides a label $c \in \mathcal{C}$, where $\mathcal{C} = \{\insider{}, \outsider{}, \textit{N/A}\}$ (hence $|\mathcal{C}|=3$). The triplets of post-phrase pairs along with their labels form the dataset $\dataset = \left\{\big((\post_i, \noun_i), c_i\big)\right\}_{i=1}^{\lvert\dataset\rvert}$. Note that a single post can appear in multiple triplets, because multiple different noun phrases can be extracted and labeled from a single post.
% TODO: histogram of the number of labels per post
The overall class distribution and a few conditional class distributions across the labeled samples for several particular noun phrases are provided in Figure \ref{fig:histogram} in the Appendix \ref{app:fig-tab}.

% Extracting these noun phrases per post \iffalse -- dependency parsing has built-in sentence segmentation --\fi helps create a dataset $\mathcal{D}$ , $(\mathtt{p},\mathtt{n}) \in \mathcal{D}$ where $(\mathtt{p},\mathtt{n})$ is a tuple of a post $\mathtt{p}$ and an identified noun phrase $\mathtt{n}$ in the post. For labeling, we use Amazon Mechanical Turk (AMT) (see Appendix \ref{app:AMT} for labeler instructions). For each tuple, the labeler is asked whether noun phrase $\mathtt{n}$ is perceived by the writer of the post \textit{given the context} $\mathtt{p}$ to be an  \textit{Insider}, \textit{Outsider} or N/A, $|\mathcal{C}|=3$ classes.. The total class distribution (top-left) and a few conditional class distributions across the labeled samples for particular noun phrases -- ``I'', ``they'', ``vaccine'', ``herd immunity'', and ``microchip'' -- are provided in Figure \ref{fig:histogram}. 

Manual inspection of the labeled samples $((\mathtt{p},\mathtt{n}),c)$ suggests that the quality of the dataset is good ($<10\%$ misclassified by random sampling). The now-labeled CT5K dataset \cite{our_dataset}\footnote{See: \href{https://osf.io/hgnm7/}{Data and Model Checkpoints}} ($|\mathcal{D}|=5000$ samples) is split into training ($90\%$), and $10\%$ testing sets. $10\%$ of the training set is held out for validation. The final training set is $20$-fold augmented by BERT-driven multi-token insertion \cite{nlpaug}.

\section{Methodology and Pipeline}
% The NLP task is to construct a context-aware model (NP2IO) that can classify all the noun phrases from an entire post as \textit{Insiders}, \textit{Outsiders} or N/A from the CT5K dataset. 

% NP2IO must also conform to the requirements in Section \ref{io}:
% \begin{itemize}
% \item \textbf{Pretrained Language Modeling:} Contextuality requires that Insiders and Outsiders are classified with respect to the post-level context. The pretraining in recent attention-based models provides out-of-the-box support to model this context. Transitivity is learned during the finetuning process of the classifier.
% \item \textbf{Competitive Zero-Shot Performance:} NP2IO's predictions must generalize to agents unseen during finetuning (Zero-shot Performance) thereby demonstrating its retention of sensitivity to contextual signals in language acquired during pre-training.
% \end{itemize}

% \noindent \textbf{Approach:} 

The \textbf{N}oun-\textbf{P}hrase-to-\textit{\textbf{I}nsider}-\textit{\textbf{O}utsider} (NP2IO) model \footnote{Code Repository: \href{https://github.com/pholur/situation-modeling}{NP2IO}} adopts a token classification architecture comprising a BERT-like pre-trained backbone and a softmax classifier on top of the backbone. Token-level labels are induced from the span-level labels for the fine-tuning over CT5K, and the span-level labeling of noun phrases is done through majority vote during inference.

% NP2IO builds upon a well-established BERT backbone pre-trained with Masked Language Modeling (MLM) and with a token classification head that labels noun phrase level spans by majority vote \footnote{Code Repository (released upon publication): \href{https://github.com/pholur/situation-modeling}{NP2IO}}.

\begin{figure*}
    \centering
    \includegraphics[width=\textwidth]{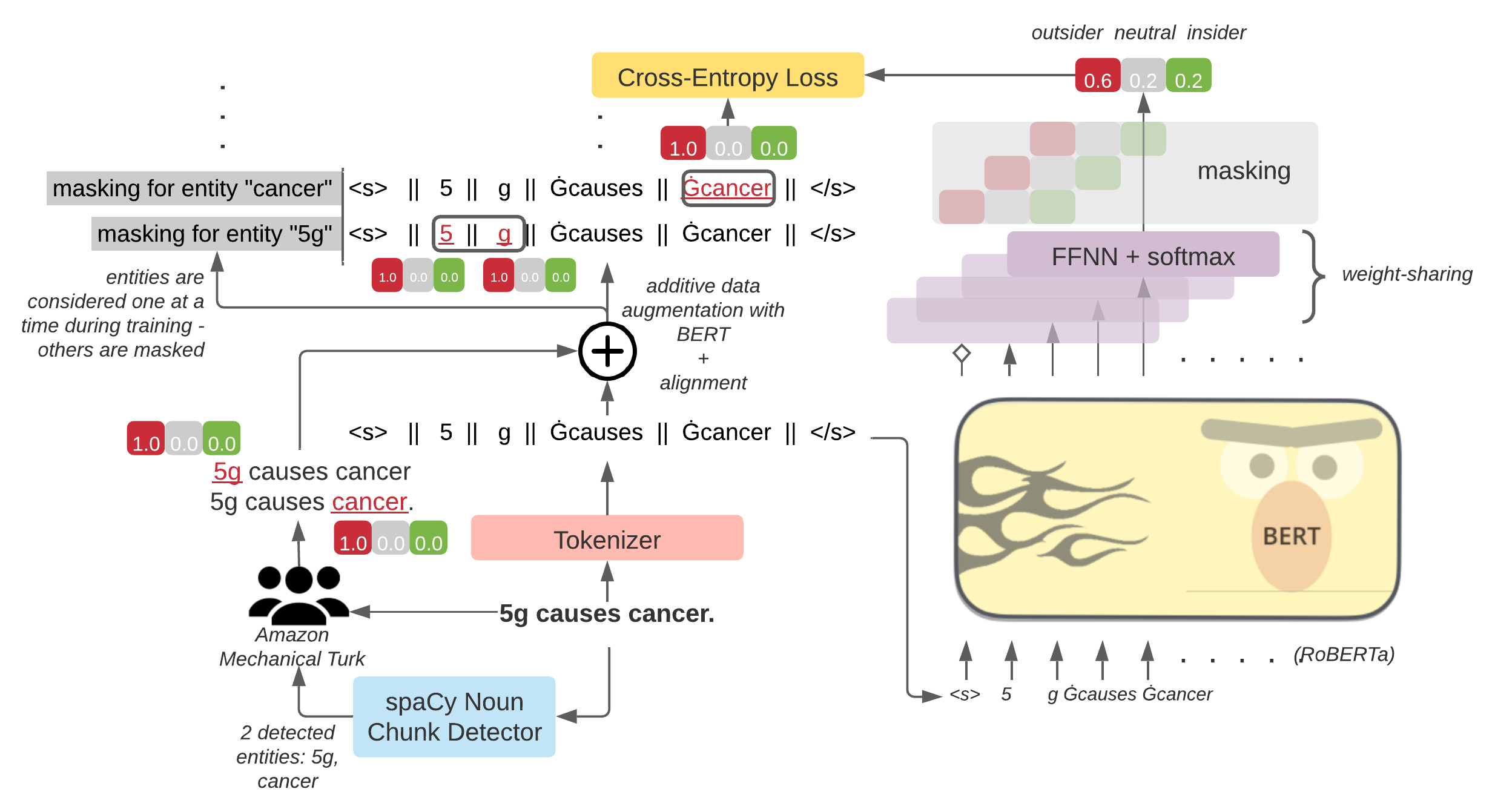}
    \caption{\textbf{NP2IO - An Outline of the Fine-tuning Pipeline:} A post is tokenized and aligned to noun chunks that are independently identified from the post with a pre-trained SpaCy parser. The BERT model is fine-tuned to identify the labels of each token in context of a post based on AMT labels of the higher-order noun phrases. Loss is Cross-Entropy (CE) loss computed on only tokens relevant for detection post SpaCy-noun phrase identification. }
    \label{training_pipeline}
\end{figure*}

\subsection{Fine-tuning Details}
\label{sec:finetune}

An outline of the fine-tuning pipeline is provided in Figure \ref{training_pipeline}.

Given a labeled example $((\post,\noun),c)$, the model labels each token $\token_i$ in the post $\post = [\token_1, \ldots, \token_N]$, where $N$ is the number of tokens in the post $\post$.
The BERT-like backbone embeds each token $\token_i$ into a contextual representation $\Phi_i \in \mathbb{R}^{d}$ (for example, $d=768$ for BERT-base or RoBERTa-base). The embedding is then passed to the softmax classification layer
\begin{align}
 \pred_i \triangleq \operatorname{Softmax}(\mathbf{W}^T\Phi_{i} + \mathbf{b}) 
\end{align}
where $\pred_i \in \Delta^{|\mathcal{C}|}$ is the \insider{}-\outsider{} classification prediction probability vector of the $i$\textsuperscript{th} token, and $\mathbf{W}\in \mathbb{R}^{d\times|\mathcal{C}|}$ and $\mathbf{b} \in \mathbb{R}^{|\mathcal{C}|}$ are the parameters of the classifier.

% The ground truth class label $c$ can be represented as a one-hot encoded vector $\mathbf{c} \in \{0,1\}^{|\mathcal{C}|}$.
The ground truth class label $c$ accounts for all occurrences of the noun phrase $\noun$ in the post $\post$. We use this span-level label to induce the token-level label and facilitate the computation of the fine-tuning loss.
% However, to facilitate the computation of the fine-tuning loss, token-level labels are needed.

Concretely, consider the spans where the noun phrase $\noun$ occurs in the post $\post$: $S_\noun = \{s_1, \ldots, s_M\}$, where $s_j \in S_\noun$ denotes the span of the $j$\textsuperscript{th} occurrence of $\noun$, and $M$ is the number of occurrences of $\noun$ in $\post$. Each span is a sequence of one or more tokens. The set of tokens appearing in one of these labeled spans is:
\begin{align}
T_\noun = \{t \in \post \;\left|\; \exists s \in S_\noun \;\; \text{s.t.} \; t \in s \right\}.
\end{align}

% The token-level labels are induced from $S_\noun$ and $\mathbf{c}$ for the tokens in T:
% \begin{align*}
%     \mathbf{c}_i = \mathbf{c} \qquad \forall t_i \in T
% \end{align*}
We define the fine-tuning loss $\loss$ of the labeled example $((\post, \noun), c)$ as the cross-entropy (CE) loss computed over $T_\noun$ using $c$ as the label for each token in it,
\begin{align}
    \loss(\post,\noun,c) = \sum_{i: t_i \in T_\noun} -\log \big(\,(\pred_i)_c\,\big)
\end{align}
where $(\pred_i)_c$ denotes the prediction probability for the class $c \in \mathcal{C}$ of the $i$\textsuperscript{th} token.

The fine-tuning is done with mini-batch gradient descent for the classification layer and a number of self-attention layers in the backbone. The number of fine-tuned self-attention layers is a hyperparameter. The scope of hyperparameter tuning is provided in Table \ref{tab:params}.

% \begin{figure*}
%     \centering
%     \includegraphics[width=0.7\textwidth]{testingpipeline.png}
%     \caption{\textbf{NP2IO - An Outline of the Testing Pipeline:} Similar to the finetuning process, a candidate test post is tokenized and an identical SpaCy noun phrase detector detects and aligns the noun phrases to their constituent tokens. Test data is evaluated only on the noun phrase tagged by AMT users via majority vote of the constituent labels of the tokens. Real-time inference as in Figure \ref{fig:intro} is attempted on all detected noun phrases .}
%     \label{fig:testing_pipeline}
% \end{figure*}

% \subsubsection{Real-time Inference and Accuracy Measurement}
\subsection{Real-time Inference and Accuracy Measurement}
During fine-tuning, we extend the label of a noun phrase to all of its constituent tokens; during inference, conversely, we summarize constituent token labels to classify the noun phrases by a majority vote. For a pair of post and noun-phrase $(\post,\noun)$, assuming the definition of $\{t_i\}_{i=1}^N$, $\{\pred_i\}_{i=1}^N$ and $T_\noun$ from the Section \ref{sec:finetune}, the \insider{}-\outsider{} label prediction $\hat{c}$ is given by
\begin{align}
\hat{c} = \argmax_{k} \sum_{i: t_i \in T_\noun} \mathbbm{1}_{\left\{k = \left(\argmax_{\kappa} (\pred_i)_\kappa\right)\right\}}.
\end{align}
% \begin{align}
% \hat{c} = \texttt{Maj.Vote} [\dots, &\argmax(\hat{\mathbf{c}}_i), \dots, \nonumber \\ & \argmax(\hat{\mathbf{c}}_j), \dots].
% \end{align}

Now $c$ can be compared to $\hat{c}$ with a number of classification evaluation metrics. Visual display of individual inference results such as those in Figure \ref{fig:intro} are supported by displaCy \cite{spacy2}. 
% Real-time graphics  such as those in Figure \ref{fig:intro} are supported by displaCy \cite{spacy2}. 

\section{Baseline Models}
In this section, we list baselines that we compare to our model's performance ordered by increasing parameter complexity.

\begin{itemize}[leftmargin=1em,topsep=0.5em,itemsep=0em,parsep=0.5em]
\item \textbf{Random Model (RND):} Given a sample from the testing set $\{\mathtt{p},\mathtt{n}\}$, $\hat{c}$ is randomly selected with uniform distribution from $\mathcal{C} = \{\insider{}, \outsider{}, \textit{N/A}\}$. 

\item \textbf{Deterministic Model (DET\;-\;I/O/NA):} For any post-phrase pair $(\post{},\noun{})$, give a fixed classification prediction: $\hat{c} = \insider{}$ (DET-I), $\hat{c} = \outsider$ (DET-O) or $\hat{c} = \textit{N/A}$ (DET-NA). 

\item \textbf{Naïve Bayes Model (NB / NB-L):} Given a training set, the naïve Bayes classifier estimates the likelihood of each class conditioned on a noun chunk $\mathcal{P}_{\mathcal{C},\mathcal{N}}(c | \mathtt{n})$ assuming its independence w.r.t. the surrounding context. That is, a noun phrase predicted more frequently in the training-set as an \insider{} will be predicted as an \insider{} during the inference, regardless of the context. For noun phrases not encountered during training, the uniform prior distribution over $\mathcal{C}$ is used for the prediction. The noun chunk may be lemmatized (by word) during training and testing to shrink the conditioned event space. We abbreviate the naïve Bayes model without lemmatization as NB, and the one with lemmatization as NB-L. 

\item \textbf{GloVe+CBOW+XGBoost (CBOW\;-\;1/2/5):} This baseline takes into account the context of a post but uses global word embeddings, instead of contextual-embeddings. A window length $w$ is fixed such that for each noun phrase, we extract the $w$ words before and $w$ words after the noun phrase, creating a set of context words, $\mathcal{S}_{w}$. Stopwords are filtered, and the remaining context words are lemmatized and encoded via $300$-dimensional GloVe \cite{glove}. The Continuous Bag of Words (CBOW) model \cite{cbow} averages the representative GloVe vectors in $\mathcal{S}_{w}$ to create an aggregate contextual vector for the noun phrase. XGBoost \cite{xgboost} is used to classify the aggregated contextual vector. The same model is applied on the test set to generate labels. We consider window lengths of $1$, $2$ and $5$ (CBOW-1, CBOW-2 and CBOW-5 respectively).

\end{itemize}

\begin{table*}[!h]
\centering
\begin{tabular}{l|lllll|lllll}
\toprule
\hline
 & \multicolumn{5}{c|}{\textbf{Performance on the Test Set}} & \multicolumn{5}{c}{\textbf{Performance in Zero-Shot}} \\ \hline
\textbf{\textbf{Model}} & \multicolumn{1}{l}{\textbf{\textbf{Acc.}}} & \multicolumn{1}{l}{\textbf{\textbf{P}}} & \multicolumn{1}{l}{\textbf{R}} & \multicolumn{1}{l}{\textbf{F1}} & \textbf{F1(w)} & \multicolumn{1}{l}{\textbf{\textbf{Acc.}}} & \multicolumn{1}{l}{\textbf{\textbf{P}}} & \multicolumn{1}{l}{\textbf{R}} & \multicolumn{1}{l}{\textbf{F1}} & \textbf{F1(w)} \\ \hline
RND & \multicolumn{1}{l}{0.334} & \multicolumn{1}{l}{0.343} & \multicolumn{1}{l}{0.334} & \multicolumn{1}{l}{0.321} & 0.350 & \multicolumn{1}{l}{0.280} & \multicolumn{1}{l}{0.273} & \multicolumn{1}{l}{0.241} & \multicolumn{1}{l}{0.239} & 0.316 \\ \hline
DET-I & \multicolumn{1}{l}{0.312} & \multicolumn{1}{l}{0.104} & \multicolumn{1}{l}{0.333} & \multicolumn{1}{l}{0.159} & 0.148 & \multicolumn{1}{l}{0.280} & \multicolumn{1}{l}{0.093} & \multicolumn{1}{l}{0.333} & \multicolumn{1}{l}{0.146} & 0.123 \\ \hline
DET-O & \multicolumn{1}{l}{0.504} & \multicolumn{1}{l}{0.168} & \multicolumn{1}{l}{0.333} & \multicolumn{1}{l}{0.223} & 0.338 & \multicolumn{1}{l}{0.593} & \multicolumn{1}{l}{0.198} & \multicolumn{1}{l}{0.333} & \multicolumn{1}{l}{0.248} & 0.442 \\ \hline
DET-NA & \multicolumn{1}{l}{0.184} & \multicolumn{1}{l}{0.061} & \multicolumn{1}{l}{0.333} & \multicolumn{1}{l}{0.104} & 0.057 & \multicolumn{1}{l}{0.127} & \multicolumn{1}{l}{0.042} & \multicolumn{1}{l}{0.333} & \multicolumn{1}{l}{0.075} & 0.028 \\ \hline
NB & \multicolumn{1}{l}{0.520} & \multicolumn{1}{l}{0.473} & \multicolumn{1}{l}{0.478} & \multicolumn{1}{l}{0.474} & 0.523 & \multicolumn{1}{l}{0.333} & \multicolumn{1}{l}{0.341} & \multicolumn{1}{l}{0.310} & \multicolumn{1}{l}{0.295} & 0.369 \\ \hline
NB-L & \multicolumn{1}{l}{0.468} & \multicolumn{1}{l}{0.397} & \multicolumn{1}{l}{0.387} & \multicolumn{1}{l}{0.386} & 0.453 & \multicolumn{1}{l}{0.360} & \multicolumn{1}{l}{0.389} & \multicolumn{1}{l}{0.434} & \multicolumn{1}{l}{0.356} & 0.373 \\ \hline
CBOW-1 & \multicolumn{1}{l}{0.490} & \multicolumn{1}{l}{0.419} & \multicolumn{1}{l}{0.383} & \multicolumn{1}{l}{0.373} & 0.448 & \multicolumn{1}{l}{0.527} & \multicolumn{1}{l}{0.408} & \multicolumn{1}{l}{0.361} & \multicolumn{1}{l}{0.360} & 0.489 \\ \hline
CBOW-2 & \multicolumn{1}{l}{0.520} & \multicolumn{1}{l}{0.462} & \multicolumn{1}{l}{0.415} & \multicolumn{1}{l}{0.410} & 0.484 & \multicolumn{1}{l}{0.553} & \multicolumn{1}{l}{0.441} & \multicolumn{1}{l}{0.375} & \multicolumn{1}{l}{0.368} & 0.509 \\ \hline
CBOW-5 & \multicolumn{1}{l}{0.526} & \multicolumn{1}{l}{0.459} & \multicolumn{1}{l}{0.419} & \multicolumn{1}{l}{0.414} & 0.489 & \multicolumn{1}{l}{0.553} & \multicolumn{1}{l}{0.393} & \multicolumn{1}{l}{0.375} & \multicolumn{1}{l}{0.369} & 0.514 \\ \hline
\textbf{NP2IO} & \multicolumn{1}{l}{\textbf{0.650}} & \multicolumn{1}{l}{\textbf{0.629}} & \multicolumn{1}{l}{\textbf{ 0.546}} & \multicolumn{1}{l}{\textbf{0.534}} & \textbf{0.619 } & \multicolumn{1}{l}{\textbf{0.693}} & \multicolumn{1}{l}{\textbf{0.682}} & \multicolumn{1}{l}{\textbf{0.536}} & \multicolumn{1}{l}{\textbf{0.543}} & \textbf{0.671} \\ \hline
\bottomrule
\end{tabular}
\caption{\textbf{Performance of NP2IO versus multiple baselines on the test set:} Our model (in bold) performs competitively and outperforms Naïve Bayes, CBOW models across metrics. Furthermore, it retains its performance to classify noun phrases unseen (post-lemmatization and stopword removal) during training. Predictably, the performance of the Naïve Bayes classifier in this zero-shot setting drops drastically to near random.}
\label{tab:full_res}
\end{table*}

\vspace{-0.4em}
\section{Results and Evaluation}
Comparison of NP2IO to baselines is provided in Table \ref{tab:full_res}. The random (RND) and deterministic (DET-I, DET-O, DET-NA) models perform  poorly. We present these results to get a better sense of the unbalanced nature of the labels in the CT5K dataset (see Figure \ref{fig:histogram}). The naïve Bayes model (NB) and its lemmatized form (NB-L) outperform the trivial baselines. However, they perform \textit{worse} than the two contextual models, GloVe+CBOW+XGBoost and NP2IO. This fact validates a crucial property of our dataset: \textit{Despite the bias in the gold standard labels for particular noun phrases such as ``I'',``they'' and ``microchip'' -- see Figure \ref{fig:histogram} in Appendix~\ref{app:fig-tab} -- context dependence plays a crucial role in \insider{}-\outsider{} classification.} Furthermore, NP2IO outperforms GloVe+CBOW+XGBoost (CBOW-1, CBOW-2, CBOW-5) summarily. \iffalse, implying that w\fi While both types of models employ context-dependence to classify noun phrases, NP2IO does so more effectively. The fine-tuning loss convergence plot for the optimal performing NP2IO model is presented in Figure~\ref{fig:convergence} in Appendix~\ref{app:fig-tab} and model checkpoints are uploaded in the data repository.

\begin{table}
\centering
\begin{tabular}{@{}l|ccc@{}}
\toprule
\hline
\textbf{Test Dataset}            & \multicolumn{3}{c}{\textbf{Train Dataset}} \\ \hline
\multicolumn{1}{l|}{}            & Laptop   & Restaurants & Tweets   \\ \hline
\multicolumn{1}{l|}{\hyperlink{cite.pontiki-etal-2014-semeval}{Laptop}}      & 0.804 & 0.768    & 0.658 \\ \hline
\multicolumn{1}{l|}{\hyperlink{cite.pontiki-etal-2014-semeval}{Restaurants}} & 0.754 & 0.825    & 0.657 \\ \hline
\multicolumn{1}{l|}{\hyperlink{cite.dong-etal-2014-adaptive}{Tweets}}      & 0.526 & 0.546    & 0.745 \\ \hline
\multicolumn{1}{l|}{\textbf{CT5K}}        & \textbf{0.347} & \textbf{0.424}   & \textbf{0.412}\\ 
\hline
\bottomrule
\end{tabular}
\caption{\textbf{F1-macro scores for the ABSA model trained on conventional SA datasets from SemEval 2014 task 4:} All models perform poorly in testing on the CT5K dataset while performing well in testing on ABSA datasets. This suggests that the CT5K dataset is indeed differentiated from the ABSA datasets.}
\label{tab:absa}
\end{table}
% with the help of BERT token representations that are, by definition, contextual in nature. This is in contrast to the GloVe+CBOW+XGBoost baseline wherein context word vector representations are aggregated as a proxy representation for context. 

% In such a model, we are not able to model the semantic structure of a conspiracy theoretic post as seamlessly as BERT's architecture in NP2IO.

\subsection{Does CT5K really differ from prior ABSA datasets?} \label{tianyi_og}

Given the limitations of current ABSA datasets for our task (see Section~\ref{io} and Section~\ref{related}), we computationally show that CT5K is indeed a different dataset, particularly in comparison to other classical ones in Table~\ref{tab:absa}. For this experiment, we train near-state-of-the-art ABSA models with RoBERTa-base backbone \cite{absa_rel} on three popular ABSA datasets -- Laptop reviews and Restaurant reviews from SemEval 2014 task 4 \cite{pontiki-etal-2014-semeval}, and Tweets \cite{dong-etal-2014-adaptive}. Each trained model is then evaluated on all three datasets \textit{as well as} the test set of CT5K. The \insider{} class in CT5K is mapped to the \textit{positive} sentiment and the \outsider{} class to the \textit{negative} sentiment. The F1-macro scores of the models trained and tested among the three ABSA datasets are much higher than the scores when testing on the CT5K dataset. \textit{Clearly, models that are successful with typical ABSA datasets do not effectively generalize to CT5K, suggesting that our dataset is different.}

\subsection{Classifying Noun Phrases at Zero-shot}

A challenge for any model, such as NP2IO, is zero-shot performance, when it encounters noun phrases never tagged during training. Answering this question offers a means for validating the context-dependence requirement, mentioned in Section~\ref{io}. This evaluation is conducted on a subset of the entire testing set: A sample of the subset $\{\mathtt{p},\mathtt{n}\}$ is such that the word-lemmatized, stopword-removed form of $\mathtt{n}$ does not exist in the set of word-lemmatized, stopword-removed noun phrases seen during training. We extract $30\%$ of test samples to be in this set. The results are presented in Table~\ref{tab:full_res}. As expected, the performance of the naïve Bayes models (NB, NB-L) degrades severely to random. The performance of the contextual models CBOW-1/2/5, and NP2IO stay strong, suggesting effective context sensitivity in inferring the correct labels for these models. A visualization of the zero-shot capabilities of NP2IO on unseen noun phrases is presented in Figure~\ref{fig:heatmap1} in Appendix~\ref{app:fig-tab}.

\begin{figure*}
    \centering
    \includegraphics[width=0.8\textwidth]{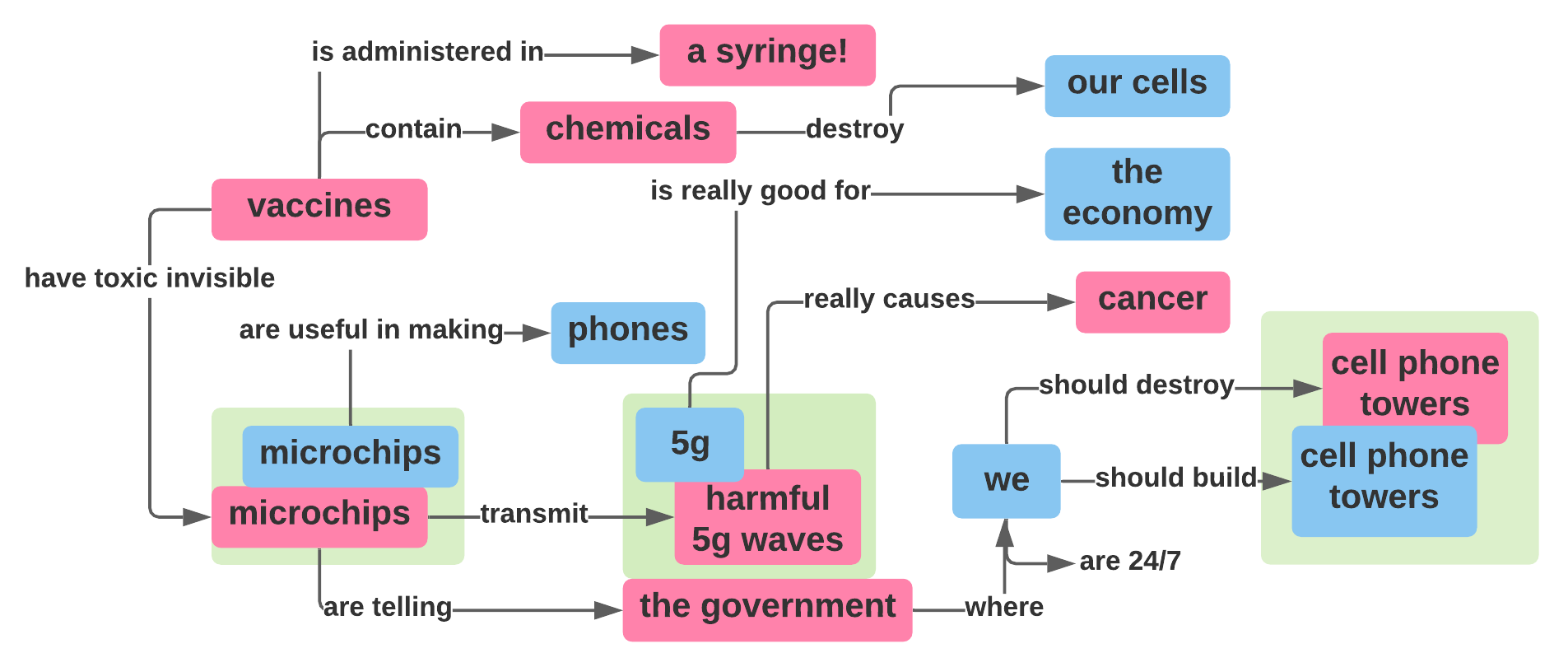}
    \caption{\textbf{An actor-actant subnarrative network constructed from social media posts:}  Selected posts from anti-vaccination forums such as  \textit{qresearch} on 4chan were decomposed into relationship tuples using a state-of-the-art relationship extraction pipeline from previous work \cite{tangherlini_2} and these relationships are overlayed with the inferences from NP2IO. This results in a network where the nodes are the noun phrases and the edges are the verb phrases, with each edge representing an extracted relationship from a post. In this network, a connected component emerged capturing a major sub-theory in vaccine hesitancy. This highlights NP2IO's ability at inferring both the threat-centric orientation of the narrative space as well as the negotiation dynamics in play, thereby providing qualitative insight into how NP2IO may be used in future work to extract large-scale relationship networks that are interpretable. The green boxes highlight the noun phrases that have contradictory membership in the \insider{}s and the \outsider{}s classes as their affiliations are deliberated.}
    \label{fig:examples}
\end{figure*}
\subsection{Does NP2IO Memorize? An Adversarial Experiment}
% Even though the gold-standard labels for a particular noun phrase in CT5K may skew toward a particular class across posts -- for example, in Figure~\ref{fig:histogram}, ``microchips'' are only labeled as \outsider{}s -- we have already quantitatively demonstrated that NP2IO remains resistant to data bias due to pre-training (exceeding performance of NB and NB-L classifiers).
We construct a set of adversarial samples to evaluate the extent to which NP2IO accurately classifies a noun phrase that has a highly-biased label distribution in CT5K. We consider $3$ noun phrases in particular: ``microchip'', ``government'', and ``chemical''. Each of these has been largely labeled as \outsider{}s. The adversarial samples for each phrase, in contrast, are manually aggregated ($5$ seed posts augmented $20$ times each) to suggest that the phrase is an \insider{} (see Table~\ref{tab:adv_base} in Appendix~\ref{app:fig-tab} for the seed posts). We compute the recall of NP2IO in detecting these \insider{} labels (results in Table~\ref{tab:adv}). NP2IO is moderately robust against adversarial attacks: \textit{In other words, highly-skewed distributions of labels for noun phrases in our dataset do not appear to imbue a similar drastic bias into our model.}

\begin{table}
    \centering
\adjustbox{max width=\columnwidth}{%
\begin{tabular}{l|ll}
\toprule
\hline
\textbf{\begin{tabular}[c]{@{}l@{}}
%Probed Noun\\
Noun \\Phrases\end{tabular}} & \textbf{\begin{tabular}[c]{@{}l@{}}CT5K \outsider{}\\ Labels (\%)\end{tabular}} & \textbf{\begin{tabular}[c]{@{}l@{}} \insider{} Recall in \\ adversarial text\end{tabular}} \\ \hline
microchip & 100\% & \textbf{80\%} \\ \hline
government & 80\% & \textbf{89\%} \\ \hline
chemical & 100\% & \textbf{62\%} \\ \hline
\bottomrule
\end{tabular}}
    \caption{\textbf{Adversarial inferencing tasks for the trained NP2IO model:} Three noun phrases with very high \outsider{} status (100\%, 80\%, 100\%, respectively) in the CT5K training set are used to construct posts where their contextual role is beneficial, and hence, should be labeled as \insider{} (see Section \ref{io}). The results show that NP2IO largely learned to use the contextual information for its inference logic, and did not memorize the agent bias in CT5K. We speculate that  the exhibited bias towards ``Chemicals'' is due to the large body of text documents that discusses the adverse effects of chemicals, and hence is encoded in the embedding structure of pretrained LM models that NP2IO cannot always overrule; at least yet. 
    %Performance on posts that contain adversarial \insider{}-labeled noun phrases having a high fraction of \outsider{} labels in our dataset:} The recall is capturing the number of instances where \insider{} labels were correctly awarded in the adversarial set.
    }
    \label{tab:adv}
    \vspace{-0.5em}
\end{table}

% Results in Table~\ref{tab:adv} suggest that NP2IO is moderately robust against adversarial attacks: In other words, highly-skewed distributions of labels for noun phrases in our dataset, does not seem to directly reflect a similar bias in our model.

\section{Concluding Remarks}
We presented a challenging \insider{}-\outsider{} classification task, a novel framework necessary for addressing burgeoning misinformation and the proliferation of threat narratives on social media. We compiled a labeled CT5K dataset of conspiracy-theoretic posts from multiple social media platforms and presented a competitive NP2IO model that outperforms non-trivial baselines. We have demonstrated that NP2IO is contextual and transitive via its zero-shot performance, adversarial studies and qualitative studies. We have also shown that the CT5K dataset consists of underlying information that is different from existing ABSA datasets.

Given NP2IO's ability to identify \insider{}s and \outsider{}s in a text segment, we can extend the inference engine to an entire set of interrelated samples in order to extract, visualize and \textit{interpret} the underlying narrative (see Figure~\ref{fig:examples}). This marks a first and significant step in teasing out narratives from fragmentary social media records, with many of its essential semantic parts -- such as, \insider{}/\outsider{} -- tagged in an automated fashion. As extensive evaluations of the NP2IO model show, our engine has learned the causal phrases used to designate the labels. We believe an immediate future work can identify such causal phrases, yet another step toward semantic understanding of the parts of a narrative. Broadly, work similar to this promises to expedite the development of models that rely on a computational foundation of structured information, and that are better at \textit{explaining} causal chains of inference, a particularly important feature in the tackling of misinformation. Indeed, NP2IO's success has answered the question: ``Which side are you on?'' What remains to be synthesized from language is: ``Why?''

% NP2IO, a narrative theory-inspired phrase-level \textit{Insider}-\textit{Outsider} classification model trained specifically to classify noun phrases. We also introduced several baselines for comparison with respect to a novel dataset CT5K, a corpus of 5000 labeled social media posts extracted from the echo-chamber of anti-vaccination discussions. The abilities displayed by NP2IO promise to provide a refreshed approach to current frameworks for threat classification in natural language, especially in fragmented and noisy text as available on social media. The model is contextualized and, as a result, shows clear generalizing power. Work along these lines promises immediate advances related to better moderation of social media. It also promises future advances in the field of Situation Modeling, an NLU task: decomposing posts into their narrative parts -- as we do here in a small setting into \textit{Insiders} and \textit{Outsiders} -- can better equip models to make informed decisions not simply at a word- or phrase- level (spanned by a window-length) but rather at a consensus-level. This can be accomplished on emerging narratives across social media platforms, thereby creating a semantically-informed language engine that goes beyond co-occurrence. 

\bibliography{anthology,custom}

\begin{thebibliography}{48}
\expandafter\ifx\csname natexlab\endcsname\relax\def\natexlab#1{#1}\fi

\bibitem[{Bailey(1999)}]{bailey}
Paul Bailey. 1999.
\newblock Searching for storiness: Story-generation from a reader’s
  perspective.
\newblock In \emph{Working notes of the Narrative Intelligence Symposium},
  pages 157--164.

\bibitem[{Bandari et~al.(2017)Bandari, Zhou, Qian, Tangherlini, and
  Roychowdhury}]{roja}
Roja Bandari, Zicong Zhou, Hai Qian, Timothy~R. Tangherlini, and Vwani~P.
  Roychowdhury. 2017.
\newblock \href {https://doi.org/10.1109/MC.2017.4041354} {A resistant strain:
  Revealing the online grassroots rise of the antivaccination movement}.
\newblock \emph{Computer}, 50(11):60--67.

\bibitem[{Bar-Haim et~al.(2017)Bar-Haim, Bhattacharya, Dinuzzo, Saha, and
  Slonim}]{redone1}
Roy Bar-Haim, Indrajit Bhattacharya, Francesco Dinuzzo, Amrita Saha, and Noam
  Slonim. 2017.
\newblock \href {https://aclanthology.org/E17-1024} {Stance classification of
  context-dependent claims}.
\newblock In \emph{Proceedings of the 15th Conference of the {E}uropean Chapter
  of the Association for Computational Linguistics: Volume 1, Long Papers},
  pages 251--261, Valencia, Spain. Association for Computational Linguistics.

\bibitem[{Barkun(2013)}]{Barkun+2013}
Michael Barkun. 2013.
\newblock \href {https://doi.org/doi:10.1525/9780520956520} {\emph{A Culture of
  Conspiracy: Apocalyptic Visions in Contemporary America}}.
\newblock University of California Press.

\bibitem[{Beatty(2016)}]{good_for}
John Beatty. 2016.
\newblock What are narratives good for?
\newblock \emph{Studies in History and Philosophy of Science Part C: Studies in
  History and Philosophy of Biological and Biomedical Sciences}, 58:33--40.

\bibitem[{Bodner et~al.(2020)Bodner, Welch, and Brodie}]{redo-bodner2020covid}
John Bodner, Wendy Welch, and Ian Brodie. 2020.
\newblock \emph{COVID-19 conspiracy theories: QAnon, 5G, the New World Order
  and other viral ideas}.
\newblock McFarland.

\bibitem[{Burki(2020)}]{burki}
Talha Burki. 2020.
\newblock \href {https://doi.org/https://doi.org/10.1016/S2589-7500(20)30227-2}
  {The online anti-vaccine movement in the age of covid-19}.
\newblock \emph{The Lancet Digital Health}, 2(10):e504--e505.

\bibitem[{Chen and Guestrin(2016)}]{xgboost}
Tianqi Chen and Carlos Guestrin. 2016.
\newblock \href {https://doi.org/10.1145/2939672.2939785} {{XGBoost}: A
  scalable tree boosting system}.
\newblock In \emph{Proceedings of the 22nd ACM SIGKDD International Conference
  on Knowledge Discovery and Data Mining}, KDD '16, pages 785--794, New York,
  NY, USA. ACM.

\bibitem[{Chong et~al.(2021)Chong, Lee, Fan, Holur, Shahsavari, Tangherlini,
  and Roychowdhury}]{dchong}
David Chong, Erl Lee, Matthew Fan, Pavan Holur, Shadi Shahsavari, Timothy
  Tangherlini, and Vwani Roychowdhury. 2021.
\newblock A real-time platform for contextualized conspiracy theory analysis.
\newblock In \emph{2021 International Conference on Data Mining Workshops
  (ICDMW) (forthcoming)}. IEEE.

\bibitem[{Clover(1986)}]{clover1986long}
Carol~J Clover. 1986.
\newblock The long prose form.
\newblock \emph{Arkiv f{\"o}r nordisk filologi}, 101:10--39.

\bibitem[{Dai et~al.(2021)Dai, Yan, Sun, Liu, and Qiu}]{absa_rel}
Junqi Dai, Hang Yan, Tianxiang Sun, Pengfei Liu, and Xipeng Qiu. 2021.
\newblock \href {https://doi.org/10.18653/v1/2021.naacl-main.146} {Does syntax
  matter? {A} strong baseline for aspect-based sentiment analysis with
  roberta}.
\newblock In \emph{Proceedings of the 2021 Conference of the North American
  Chapter of the Association for Computational Linguistics: Human Language
  Technologies, {NAACL-HLT} 2021, Online, June 6-11, 2021}, pages 1816--1829.
  Association for Computational Linguistics.

\bibitem[{Dong et~al.(2014)Dong, Wei, Tan, Tang, Zhou, and
  Xu}]{dong-etal-2014-adaptive}
Li~Dong, Furu Wei, Chuanqi Tan, Duyu Tang, Ming Zhou, and Ke~Xu. 2014.
\newblock \href {https://doi.org/10.3115/v1/P14-2009} {Adaptive recursive
  neural network for target-dependent {T}witter sentiment classification}.
\newblock In \emph{Proceedings of the 52nd Annual Meeting of the Association
  for Computational Linguistics (Volume 2: Short Papers)}, pages 49--54,
  Baltimore, Maryland. Association for Computational Linguistics.

\bibitem[{Du et~al.(2017)Du, Xu, He, and Gui}]{redone2}
Jiachen Du, Ruifeng Xu, Yulan He, and Lin Gui. 2017.
\newblock \href {https://doi.org/10.24963/ijcai.2017/557} {Stance
  classification with target-specific neural attention}.
\newblock In \emph{Proceedings of the Twenty-Sixth International Joint
  Conference on Artificial Intelligence, {IJCAI-17}}, pages 3988--3994.

\bibitem[{Dundes(1962)}]{dundes1962morphology}
Alan Dundes. 1962.
\newblock \emph{The Morphology of North American Indian Folktales}.
\newblock Indiana University.

\bibitem[{Gao et~al.(2021)Gao, Wang, Liu, Wang, Zhang, and Liao}]{aspect}
Lei Gao, Yulong Wang, Tongcun Liu, Jingyu Wang, Lei Zhang, and Jianxin Liao.
  2021.
\newblock \href {https://ojs.aaai.org/index.php/AAAI/article/view/17523}
  {Question-driven span labeling model for aspect–opinion pair extraction}.
\newblock \emph{Proceedings of the AAAI Conference on Artificial Intelligence},
  35(14):12875--12883.

\bibitem[{Holur et~al.(2022)Holur, Chong, Lee, Fan, Shahsavari, Wang,
  Tangherlini, and Roychowdhury}]{our_dataset}
Pavan Holur, David Chong, Erl Lee, Matthew Fan, Shadi Shahsavari, Tianyi Wang,
  Timothy~R Tangherlini, and Vwani Roychowdhury. 2022.
\newblock \href {osf.io/hgnm7} {Acl 2022 - supplementary data files}.

\bibitem[{Holur et~al.(2021)Holur, Shahsavari, Ebrahimzadeh, Tangherlini, and
  Roychowdhury}]{rsos}
Pavan Holur, Shadi Shahsavari, Ehsan Ebrahimzadeh, Timothy~R. Tangherlini, and
  Vwani Roychowdhury. 2021.
\newblock \href {https://doi.org/10.1098/rsos.210797} {Modelling social
  readers: novel tools for addressing reception from online book reviews}.
\newblock \emph{Royal Society Open Science}, 8(12):210797.

\bibitem[{Honnibal and Johnson(2015)}]{spacy_OG}
Matthew Honnibal and Mark Johnson. 2015.
\newblock \href {https://doi.org/10.18653/v1/D15-1162} {An improved
  non-monotonic transition system for dependency parsing}.
\newblock In \emph{Proceedings of the 2015 Conference on Empirical Methods in
  Natural Language Processing}, pages 1373--1378, Lisbon, Portugal. Association
  for Computational Linguistics.

\bibitem[{Honnibal and Montani(2017)}]{spacy2}
Matthew Honnibal and Ines Montani. 2017.
\newblock {spaCy 2}: Natural language understanding with {B}loom embeddings,
  convolutional neural networks and incremental parsing.
\newblock To appear.

\bibitem[{Kandias et~al.(2013)Kandias, Stavrou, Bozovic, and
  Gritzalis}]{kandias}
Miltiadis Kandias, Vasilis Stavrou, Nick Bozovic, and Dimitris Gritzalis. 2013.
\newblock Proactive insider threat detection through social media: The youtube
  case.
\newblock In \emph{Proceedings of the 12th ACM workshop on Workshop on privacy
  in the electronic society}, pages 261--266.

\bibitem[{K{\"u}{\c c}{\"u}k and Can(2021)}]{kucukStanceDetectionSurvey2021}
Dilek K{\"u}{\c c}{\"u}k and Fazli Can. 2021.
\newblock \href {https://doi.org/10.1145/3369026} {Stance {{Detection}}: {{A
  Survey}}}.
\newblock \emph{ACM Computing Surveys}, 53(1):1--37.

\bibitem[{Labov and Waletzky(1967)}]{labov1967narrative}
William Labov and Joshua Waletzky. 1967.
\newblock Narrative analysis. inj. helm (ed.), essays on the verbal and visual
  arts (pp. 12--44).

\bibitem[{Lample and Conneau(2019)}]{xlm}
Guillaume Lample and Alexis Conneau. 2019.
\newblock Cross-lingual language model pretraining.
\newblock \emph{Advances in Neural Information Processing Systems (NeurIPS)}.

\bibitem[{Li et~al.(2019)Li, Bing, Zhang, and Lam}]{aspect_2}
Xin Li, Lidong Bing, Wenxuan Zhang, and Wai Lam. 2019.
\newblock \href {https://doi.org/10.18653/v1/D19-5505} {Exploiting {BERT} for
  end-to-end aspect-based sentiment analysis}.
\newblock In \emph{Proceedings of the 5th Workshop on Noisy User-generated Text
  (W-NUT 2019)}, pages 34--41, Hong Kong, China. Association for Computational
  Linguistics.

\bibitem[{Liu et~al.(2019)Liu, Ott, Goyal, Du, Joshi, Chen, Levy, Lewis,
  Zettlemoyer, and Stoyanov}]{roberta}
Yinhan Liu, Myle Ott, Naman Goyal, Jingfei Du, Mandar Joshi, Danqi Chen, Omer
  Levy, Mike Lewis, Luke Zettlemoyer, and Veselin Stoyanov. 2019.
\newblock \href {http://arxiv.org/abs/1907.11692} {Roberta: A robustly
  optimized bert pretraining approach}.

\bibitem[{Ma(2019)}]{nlpaug}
Edward Ma. 2019.
\newblock Nlp augmentation.
\newblock https://github.com/makcedward/nlpaug.

\bibitem[{Mikolov et~al.(2013)Mikolov, Chen, Corrado, and Dean}]{cbow}
Tomas Mikolov, Kai Chen, Greg Corrado, and Jeffrey Dean. 2013.
\newblock Efficient estimation of word representations in vector space.
\newblock \emph{arXiv preprint arXiv:1301.3781}.

\bibitem[{Nicolaisen(1987)}]{nicolaisen1987linguistic}
Wilhelm~FH Nicolaisen. 1987.
\newblock The linguistic structure of legends.
\newblock \emph{Perspectives on Contemporary Legend}, 2(1):61--67.

\bibitem[{Park et~al.(2018)Park, You, and Lee}]{park}
Won Park, Youngin You, and Kyungho Lee. 2018.
\newblock Detecting potential insider threat: Analyzing insiders’ sentiment
  exposed in social media.
\newblock \emph{Security and Communication Networks}, 2018.

\bibitem[{Pennington et~al.(2014)Pennington, Socher, and Manning}]{glove}
Jeffrey Pennington, Richard Socher, and Christopher~D. Manning. 2014.
\newblock \href {http://www.aclweb.org/anthology/D14-1162} {Glove: Global
  vectors for word representation}.
\newblock In \emph{Empirical Methods in Natural Language Processing (EMNLP)},
  pages 1532--1543.

\bibitem[{Pontiki et~al.(2014)Pontiki, Galanis, Pavlopoulos, Papageorgiou,
  Androutsopoulos, and Manandhar}]{pontiki-etal-2014-semeval}
Maria Pontiki, Dimitris Galanis, John Pavlopoulos, Harris Papageorgiou, Ion
  Androutsopoulos, and Suresh Manandhar. 2014.
\newblock \href {https://doi.org/10.3115/v1/S14-2004} {{S}em{E}val-2014 task 4:
  Aspect based sentiment analysis}.
\newblock In \emph{Proceedings of the 8th International Workshop on Semantic
  Evaluation ({S}em{E}val 2014)}, pages 27--35, Dublin, Ireland. Association
  for Computational Linguistics.

\bibitem[{Sanh et~al.(2020)Sanh, Debut, Chaumond, and Wolf}]{distilbert}
Victor Sanh, Lysandre Debut, Julien Chaumond, and Thomas Wolf. 2020.
\newblock \href {http://arxiv.org/abs/1910.01108} {Distilbert, a distilled
  version of bert: smaller, faster, cheaper and lighter}.

\bibitem[{Shahsavari et~al.(2020{\natexlab{a}})Shahsavari, Ebrahimzadeh,
  Shahbazi, Falahi, Holur, Bandari, R.~Tangherlini, and Roychowdhury}]{shadi_0}
Shadi Shahsavari, Ehsan Ebrahimzadeh, Behnam Shahbazi, Misagh Falahi, Pavan
  Holur, Roja Bandari, Timothy R.~Tangherlini, and Vwani Roychowdhury.
  2020{\natexlab{a}}.
\newblock \href {https://doi.org/10.1145/3394231.3397918} {An automated
  pipeline for character and relationship extraction from readers literary book
  reviews on goodreads.com}.
\newblock In \emph{12th ACM Conference on Web Science}, WebSci '20, page
  277–286, New York, NY, USA. Association for Computing Machinery.

\bibitem[{Shahsavari et~al.(2020{\natexlab{b}})Shahsavari, Holur, Wang,
  Tangherlini, and Roychowdhury}]{tangherlini_1}
Shadi Shahsavari, Pavan Holur, Tianyi Wang, Timothy~R Tangherlini, and Vwani
  Roychowdhury. 2020{\natexlab{b}}.
\newblock Conspiracy in the time of corona: automatic detection of emerging
  covid-19 conspiracy theories in social media and the news.
\newblock \emph{Journal of computational social science}, 3(2):279--317.

\bibitem[{Socher et~al.(2013)Socher, Perelygin, Wu, Chuang, Manning, Ng, and
  Potts}]{sst}
Richard Socher, Alex Perelygin, Jean Wu, Jason Chuang, Christopher~D. Manning,
  Andrew Ng, and Christopher Potts. 2013.
\newblock \href {https://www.aclweb.org/anthology/D13-1170} {Recursive deep
  models for semantic compositionality over a sentiment treebank}.
\newblock In \emph{Proceedings of the 2013 Conference on Empirical Methods in
  Natural Language Processing}, pages 1631--1642, Seattle, Washington, USA.
  Association for Computational Linguistics.

\bibitem[{Tajfel(1974)}]{tajfel_2}
Henri Tajfel. 1974.
\newblock Social identity and intergroup behaviour.
\newblock \emph{Social science information}, 13(2):65--93.

\bibitem[{Tajfel et~al.(1979)Tajfel, Turner, Austin, and Worchel}]{tajfel_1}
Henri Tajfel, John~C Turner, William~G Austin, and Stephen Worchel. 1979.
\newblock An integrative theory of intergroup conflict.
\newblock \emph{Organizational identity: A reader}, 56(65):9780203505984--16.

\bibitem[{Tangherlini(2018)}]{tangherlini2018toward}
Timothy~R Tangherlini. 2018.
\newblock Toward a generative model of legend: Pizzas, bridges, vaccines, and
  witches.
\newblock \emph{Humanities}, 7(1):1.

\bibitem[{Tangherlini et~al.(2016)Tangherlini, Roychowdhury, Glenn, Crespi,
  Bandari, Wadia, Falahi, Ebrahimzadeh, and Bastani}]{mommy}
Timothy~R Tangherlini, Vwani Roychowdhury, Beth Glenn, Catherine~M Crespi, Roja
  Bandari, Akshay Wadia, Misagh Falahi, Ehsan Ebrahimzadeh, and Roshan Bastani.
  2016.
\newblock \href {https://doi.org/10.2196/publichealth.6586} {``mommy blogs''
  and the vaccination exemption narrative: Results from a machine-learning
  approach for story aggregation on parenting social media sites}.
\newblock \emph{JMIR Public Health Surveill}, 2(2):e166.

\bibitem[{Tangherlini et~al.(2020)Tangherlini, Shahsavari, Shahbazi,
  Ebrahimzadeh, and Roychowdhury}]{tangherlini_2}
Timothy~R Tangherlini, Shadi Shahsavari, Behnam Shahbazi, Ehsan Ebrahimzadeh,
  and Vwani Roychowdhury. 2020.
\newblock An automated pipeline for the discovery of conspiracy and conspiracy
  theory narrative frameworks: Bridgegate, pizzagate and storytelling on the
  web.
\newblock \emph{PloS one}, 15(6):e0233879.

\bibitem[{Tkachenko et~al.(2020-2021)Tkachenko, Malyuk, Shevchenko, Holmanyuk,
  and Liubimov}]{LabelStudio}
Maxim Tkachenko, Mikhail Malyuk, Nikita Shevchenko, Andrey Holmanyuk, and
  Nikolai Liubimov. 2020-2021.
\newblock \href {https://github.com/heartexlabs/label-studio} {{Label Studio}:
  Data labeling software}.
\newblock Open source software available from
  https://github.com/heartexlabs/label-studio.

\bibitem[{Vlad et~al.(2019)Vlad, Tanase, Onose, and Cercel}]{bertbi}
George-Alexandru Vlad, Mircea-Adrian Tanase, Cristian Onose, and
  Dumitru-Clementin Cercel. 2019.
\newblock \href {https://doi.org/10.18653/v1/D19-5022} {Sentence-level
  propaganda detection in news articles with transfer learning and
  {BERT}-{B}i{LSTM}-capsule model}.
\newblock In \emph{Proceedings of the Second Workshop on Natural Language
  Processing for Internet Freedom: Censorship, Disinformation, and Propaganda},
  pages 148--154, Hong Kong, China. Association for Computational Linguistics.

\bibitem[{Walker et~al.(2012)Walker, Anand, Abbott, and
  Grant}]{walkerStanceClassificationUsing2012}
Marilyn Walker, Pranav Anand, Rob Abbott, and Ricky Grant. 2012.
\newblock Stance {{Classification}} using {{Dialogic Properties}} of
  {{Persuasion}}.
\newblock In \emph{Proceedings of the 2012 {{Conference}} of the {{North
  American Chapter}} of the {{Association}} for {{Computational Linguistics}}:
  {{Human Language Technologies}}}, pages 592--596, {Montr\'eal, Canada}.
  {Association for Computational Linguistics}.

\bibitem[{Wang et~al.(2021)Wang, Jiang, Bach, Wang, Huang, Huang, and
  Tu}]{wang}
Xinyu Wang, Yong Jiang, Nguyen Bach, Tao Wang, Zhongqiang Huang, Fei Huang, and
  Kewei Tu. 2021.
\newblock {{Automated Concatenation of Embeddings for Structured Prediction}}.
\newblock In \emph{{the Joint Conference of the 59th Annual Meeting of the
  Association for Computational Linguistics and the 11th International Joint
  Conference on Natural Language Processing (\textbf{ACL-IJCNLP 2021})}}.
  Association for Computational Linguistics.

\bibitem[{Wester et~al.(2016)Wester, {\O}vrelid, Velldal, and Hammer}]{wester}
Aksel Wester, Lilja {\O}vrelid, Erik Velldal, and Hugo~Lewi Hammer. 2016.
\newblock Threat detection in online discussions.
\newblock In \emph{Proceedings of the 7th Workshop on Computational Approaches
  to Subjectivity, Sentiment and Social Media Analysis}, pages 66--71.

\bibitem[{Yang et~al.(2019)Yang, Lee, Whang, Lee, and Lim}]{emotionxku}
Kisu Yang, Dongyub Lee, Taesun Whang, Seolhwa Lee, and Heuiseok Lim. 2019.
\newblock \href {http://arxiv.org/abs/1906.11565} {Emotionx-ku: Bert-max based
  contextual emotion classifier}.

\bibitem[{Yin et~al.(2020)Yin, Meng, and Chang}]{sentibert}
Da~Yin, Tao Meng, and Kai-Wei Chang. 2020.
\newblock {SentiBERT}: A transferable transformer-based architecture for
  compositional sentiment semantics.
\newblock In \emph{Proceedings of the 58th Conference of the Association for
  Computational Linguistics, {ACL} 2020, Seattle, USA}.

\bibitem[{Zhang et~al.(2017)Zhang, Qiu, Chen, Zhang, Yu, and
  Elhadad}]{zhangWeMakeChoices2017}
Shaodian Zhang, Lin Qiu, Frank Chen, Weinan Zhang, Yong Yu, and No{\'e}mie
  Elhadad. 2017.
\newblock \href {https://doi.org/10.1145/3041021.3055134} {We {{Make Choices We
  Think}} are {{Going}} to {{Save Us}}: {{Debate}} and {{Stance
  Identification}} for {{Online Breast Cancer CAM Discussions}}}.
\newblock In \emph{Proceedings of the 26th {{International Conference}} on
  {{World Wide Web Companion}}}, {{WWW}} '17 {{Companion}}, pages 1073--1081,
  {Republic and Canton of Geneva, CHE}. {International World Wide Web
  Conferences Steering Committee}.

\end{thebibliography}
\bibliographystyle{acl_natbib}

\appendix
\appendixpage

\section{Data Collection}
\subsection{Automated Crawling of Social Media} \label{app:data_collection}
A daily data collection method \cite{dchong} aggregates heterogeneous data from various social media platforms including Reddit, YouTube, 4chan and 8kun. Our implementation of this pipeline has extracted potentially conspiracy theoretic posts between March $2020$ and June $2021$. We select a subset of these posts that are relevant to vaccine hesitancy and that include (a) at least one of the words in ['vaccine', 'mrna', 'pfizer', 'moderna', 'j\&j', 'johnson', 'chip', 'pharm'] and (b) between $150$ to $700$ characters. The end-to-end data processing pipeline is \textit{uncased}.

\subsection{Instructions to AMT Labelers} \label{app:AMT}
Amazon Mechanical Turk Labelers were required to be at the Masters' level (exceeding a trust baseline provided by Amazon), were required to speak English, and were required to be residing in the United States. No personally identifying information was collected. Users were asked to create an account on a LabelStudio \cite{LabelStudio} platform to answer a set of 60-80 questions or 2 hours worth of questions. Each question included (a) A real anonymized social media post with a highlighted sentence, (b) The sentence highlighted in (a) but with the noun phrase of interest highlighted. The question prompt read: \textit{Please let us know whether the entity highlighted in bold AS PERCEIVED BY THE WRITER is a good/bad or neutral entity.}

Labelers were reminded several times via popups and other means that the labels were to be chosen with respect to the author of the post and not the labeler's inherent biases and/or political preferences.

\subsection{Ethics statement about the collected social media data}
Data was collected using verified research Application Programming Interfaces (API) provided by the social media companies for non-commercial study. In order to explore data on fringe platforms such as 4chan and 8kun where standard APIs are not available, the data was scraped using a Selenium-based crawler. All the retrieved samples were ensured to be public: the posts could be accessed by anyone on the internet without requiring explicit consent by the authors. Furthermore, we made sure to avoid using Personal Identifiable Information (PII) such as the user location, time of posting and other metadata: indeed, we hid even the specific social media platform from which a particular post was mined. The extracted text was cleaned by fixing capitalization, filtering special characters, adjusting inter-word spacing and correcting punctuation, all of which further obfuscated the identity of the author of a particular post.

% \newpage
\section{\label{app:fig-tab}Supplementary Figures and Tables}

\begin{figure}[htbp]
    \centering
    \includegraphics[width=0.8\columnwidth]{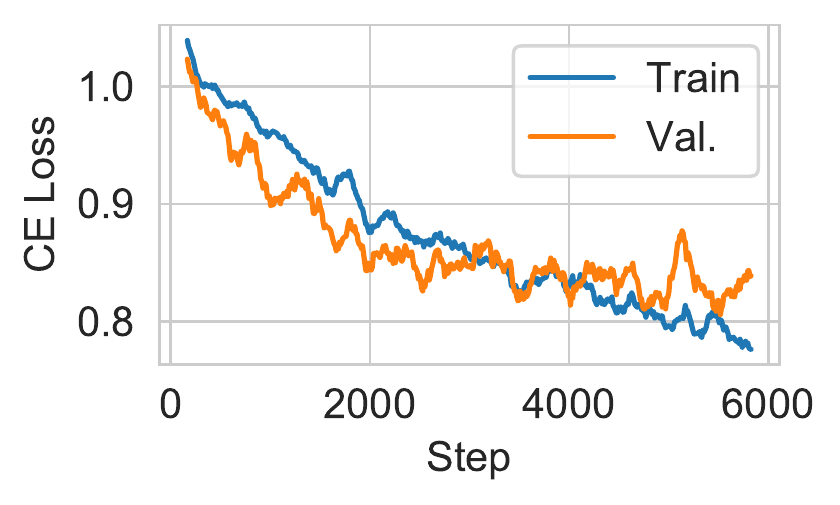}
    \caption{\textbf{Convergence plot for NP2IO:} Shown above is the training and validation CE loss with optimal parameters (in bold) from Table~\ref{tab:params}. Model checkpoints are in data repository.}
    \label{fig:convergence}
\end{figure}

\begin{table}[bp]
\centering
\begin{tabular}{l|l}
\toprule
\hline
\textbf{Parameters} & \textbf{Values} \\ \hline
Batch Size & 32, \textbf{64}, 128 \\ \hline
Trainable Layers & 0, 1, \textbf{2}, 5 \\ \hline
LR & 1E-7, \textbf{1E-6}, 1E-5, 1E-4 \\ \hline
\makecell[lc]{Pretrained\\Backbone} & \begin{tabular}[c]{@{}l@{}}BERT-base, \\\textbf{DistilBERT-base}, \\ RoBERTa-base, \\RoBERTa-large\end{tabular} \\ \hline
\bottomrule
\end{tabular}
\caption{\textbf{A summary of the parameters considered for finetuning:} NP2IO's best-performing (by validation loss) parameters are in bold.}
\label{tab:params}
\end{table}

\begin{figure}[t]
    \centering
    \includegraphics[width=1.0\columnwidth]{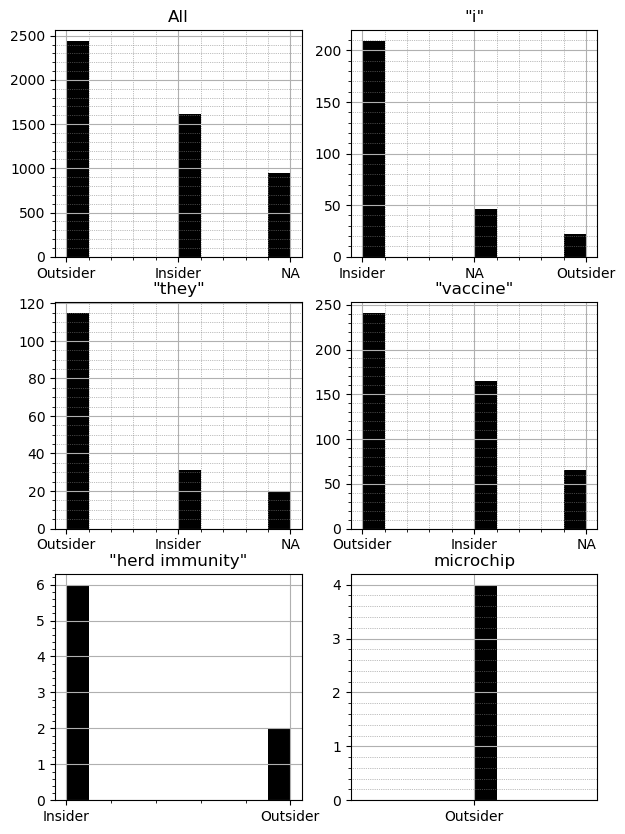}
    \caption{\textbf{Histograms that show the distributions of labels in CT5K}: The plot for ``All'' represents the full 3-category label distribution across all entities, for ``I'' the bias toward \insider{}s is evident, ``They'' are mostly outsiders, and there is no clear consensus label for ``Vaccine'' and ``Herd Immunity''. Microchips are always tagged as \outsider{}s.}
    \label{fig:histogram}
\end{figure}

% PAVAN zero-shot verification.

\begin{figure*}[b]
    \centering
    \includegraphics[width=\textwidth]{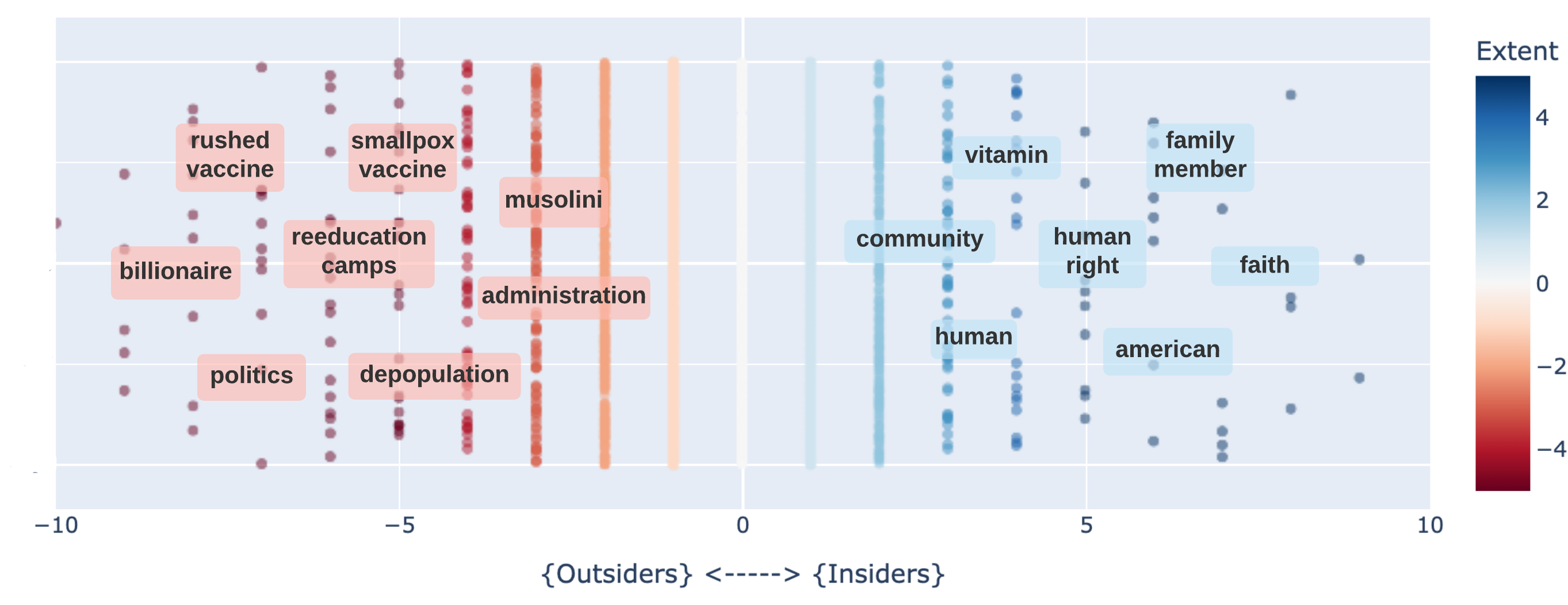}
    \caption{\textbf{Zero-shot \insider{}-\outsider{} Classification Profile:} This figure shows the consensus vote for noun phrases that do not occur in the training set. The x-axis represents the consensus-vote-count and the y-axis, the indices of the noun phrases. The consensus vote is computed for each noun phrase $\noun{}$ by passing all the posts that include $\noun$ through NP2IO. Each \textit{Insider} vote is $+1$ and \textit{Outsider} vote is $-1$. The consensus-vote-count is also color-coded for better visualization. The zero-shot classification is qualitatively observed to correctly classify popular noun phrases such as ``reeducation camps'',``depopulation'' as \outsider{}s and ``american'' and ``faith'' as \insider{}s in the subnarrative of the anti-vaccination movement.}
    \label{fig:heatmap1}
\end{figure*}

\begin{table*}
    
    \centering
\begin{tabular}{l|l}
\toprule
\hline
\textbf{NP} & \textbf{\begin{tabular}[c]{@{}l@{}}Seed Posts \\ (augmented to 100 posts per NP)\end{tabular}} \\ \hline
\textbf{microchip} & \begin{tabular}[c]{@{}l@{}}"I love \textbf{microchip}s.", "I feel that \textbf{microchip}s are great.", \\ "\textbf{microchip}s are lovely and extremely useful.", \\ "I believe \textbf{microchip}s are useful in making phones.",\\ "\textbf{Microchip}s have made me a lot of money."\end{tabular} \\ \hline
\textbf{government} & \begin{tabular}[c]{@{}l@{}}"The \textbf{government} helps keep me safe.","The \textbf{government} \\ does a good job.","I think that without the \textbf{government}, \\ we would be worse off.","The \textbf{government} keeps us safe.",\\ "A \textbf{government} is important to keep our society stable."\end{tabular} \\ \hline
\textbf{chemical} & \begin{tabular}[c]{@{}l@{}}"\textbf{Chemical}s save us.","\textbf{Chemical}s can cure cancer.","I think \\ \textbf{chemical}s can help elongate our lives.","I think \textbf{chemical}s \\ are great and helps keep us healthy.","\textbf{Chemical}s can help \\ remove ringworms."\end{tabular} \\ \hline
\bottomrule
\end{tabular}
    \caption{\textbf{The set of $5$ \insider{}-oriented core posts per noun phrase (in bold) that have a high skew toward \outsider{} labels in CT5K:} Each seed post is augmented $20$ times to create a set of $100$ adversarial posts per phrase. NP2IO infers the label for the key noun phrase across these samples. The \textit{adversarial} recall is presented in Table~\ref{tab:adv}.}
    \label{tab:adv_base}
\end{table*}

\end{document}